\documentclass[pdftex,12pt,a4paper,oneside]{article}

\usepackage{arxiv}

\usepackage[utf8]{inputenc} 
\usepackage[T1]{fontenc}    
\usepackage{hyperref}       
\usepackage{url}            
\usepackage{booktabs}       
\usepackage{amsfonts}       
\usepackage{nicefrac}       
\usepackage{microtype}      
\usepackage{lipsum}		
\usepackage{graphicx}
\usepackage{natbib}
\usepackage{doi}

\usepackage{amsmath} 
\numberwithin{equation}{section} 
\usepackage{algorithm}
\usepackage{algorithmic} 
\usepackage{fancyhdr} 
\usepackage{graphicx} 
\usepackage{lipsum}  
\usepackage{setspace} 
\usepackage[]{fncychap} 
\usepackage{tabularx}
\newcolumntype{Y}{>{\centering\arraybackslash}X}
\usepackage{appendix} 
\usepackage{natbib}
\usepackage{float}

\usepackage{booktabs}
\usepackage{makecell}
\usepackage{amssymb}
\usepackage{caption}

\hypersetup{hidelinks}
\usepackage{cleveref}

\usepackage{titlesec}

\titleformat{\section}
  {\normalfont\fontsize{18}{22}\selectfont\bfseries}
  {\thesection}{1em}{}

\titleformat{\subsection}
  {\normalfont\fontsize{15}{20}\selectfont\bfseries}
  {\thesubsection}{1em}{}

\titleformat{\subsubsection}
  {\normalfont\fontsize{13}{18}\selectfont\bfseries}
  {\thesubsubsection}{1em}{}

\title{SocraticRepair: Planning-Guided Socratic
Tutoring for Program Debugging and Misconception Repair}

\author{
    \textbf{Mykhailo Kopytskyi} \\
    University of Warwick \\
    \texttt{mykhailo.kopytskyi@warwick.ac.uk}
    \And
    \textbf{Mir Seyedebrahimi} \\
    Warwick Manufacturing Group \\
    \texttt{Mir.Seyedebrahimi@warwick.ac.uk}
    \AND
    \textbf{Jianhua Yang} \\
    Warwick Manufacturing Group \\
    \texttt{Jianhua.Yang@warwick.ac.uk}
}
\date{}

\renewcommand{\undertitle}{}
\renewcommand{\shorttitle}{SocraticRepair}

\hypersetup{
    pdftitle={SocraticRepair: Planning-Guided Socratic Tutoring for Program Debugging and Misconception Repair},
    pdfauthor={Mykhailo Kopytskyi, Mir Seyedebrahimi, Jianhua Yang},
    pdfsubject={Artificial Intelligence, Programming Education},
    pdfkeywords={Socratic tutoring, programming education, debugging, large language models}
}

\begin{document}
\raggedbottom
\maketitle

\fontsize{11}{16}\selectfont

\begin{abstract}
	Large language models are increasingly used in computer science education. However, the teaching abilities of even the most powerful LLMs are limited: they tend to reveal the solution prematurely and may lose focus on the learner's current debugging objective. To address this, we introduce SocraticRepair, a planning-guided Socratic tutoring framework for program debugging and misconception repair that adapts prior planning-guided tutoring work \citep{scaffoldLM19math}. 
The framework is instantiated in the Python programming domain by fine-tuning Qwen2.5-Coder-7B-Instruct on synthetic multi-turn tutoring dialogues constructed from four preprocessed programming datasets.
The system is evaluated using a predefined set of independent metrics and benchmarked against a general-purpose LLM. The experiments show that SocraticRepair achieves higher scores on stepwise scaffolding and question quality while also substantially reducing premature solution disclosure. The results also identify reliable pedagogical plan generation as an important remaining challenge.
\end{abstract}


\section{Introduction}
Large language models have demonstrated their usefulness in a variety of programming education scenarios, including program repair, hint generation, feedback and code explanation \citep{genProgrEducation1progr, koutchemeProgrRepairFeedback}. Their ability to generate context-sensitive responses has further motivated their use in interactive educational settings \citep{codeAid24Progr, tutorsInProgrCourse6}. Consequently, there has been growing interest in using LLMs as conversational tutors that can guide learners through problem solving. 

Designing and implementing effective tutoring remains one of the challenges.
Existing systems have explored several approaches to make student-to-tutor interactions more pedagogically aligned.
One line of work grounds general-purpose LLMs through carefully designed prompts, pedagogical guardrails and course-specific material with the aim of producing relevant assistance that avoids direct solution disclosure \citep{codeAid24Progr, tutorsInProgrCourse6}. A second line introduces more explicit structure into tutoring process by using multi-turn planning and knowledge tracing to determine how the interaction should proceed \citep{instructNotAssist23programming, traver9Progr}. More recently, supervised fine-tuning has been explored as an alternative strategy to simpler prompting-based methods \citep{ross2025supervisedfinetuningllmsbehave}. Despite the progress, prior work has mainly addressed these aspects of tutoring separately. In contrast, recent work outside programming education has demonstrated that a pedagogical tutor, grounded in a plan, can be incorporated into a specialised fine-tuned tutor \citep{scaffoldLM19math}. This leaves it underexplored whether such structured tutoring behaviour can be adapted and learned by a specialised programming model for Socratic multi-turn bug and misconception repair, where the tutor must reason not only about the learner's developing understanding, but also about the current executable state of the learner's program, whose correctness does not necessarily imply that the underlying misconception has been resolved. 

To address the gap, we introduce SocraticRepair, a planning-guided tutoring framework for Socratic multi-turn programming bug and misconception repair. Building on recent work on planning-guided tutoring \citep{scaffoldLM19math}, the approach is adapted, such that the progress involves both the learner's demonstrated understanding and the functional state of their code. SocraticRepair uses a pedagogical plan containing conceptual and repair objectives and conditions its guidance on the learner state and progress through these plan objectives. Because suitable multi-turn programming tutoring data are limited, a verified synthetic corpus is constructed from buggy student submissions drawn from four programming datasets. This corpus is used to fine-tune the Qwen2.5-Coder-7B-Instruct for planning and multi-turn tutoring. SocraticRepair is evaluated in controlled multi-turn interactions against a general-purpose LLM using 6 dimensional turn-level metrics and a binary metric of premature solution leakage.

The main contributions of this work are therefore threefold. \textbf{First}, a planning-guided multi-turn tutoring framework is adapted to Socratic multi-turn bug and misconception repair, where tutoring progress depends on both the learner's demonstrated understanding and the functional state of their code. The framework is instantiated and evaluated in the Python programming domain.
\textbf{Second}, a verified synthetic multi-turn tutoring corpus is constructed from four Python programming datasets using controlled student-to-tutor simulation and use it to fine-tune a Qwen2.5-Coder-7B-Instruct model for planning and tutoring. \textbf{Third}, the resulting tutor is evaluated under controlled interactions against a general-purpose LLM using six turn-level evaluation dimensions and a binary metric of premature solution leakage.

\section{Literature Review}

\subsection{Intelligent Tutoring Systems (ITSs)}
Before the emergence of LLM-based tutors, personalised computer-based tutoring was performed via the ITSs \citep{VanLEHN01102011,itsEffect, itsCrowProgr}. Traditional ITS architectures combined the knowledge about the target domain with a model of the learner and a pedagogical strategy for selecting the next action \citep{AndersonCognitiveTutors}. Learner knowledge was represented in different ways across the systems: model tracing compared learner actions against a cognitive model, knowledge tracing estimated the probability that individual skills had been mastered from the observed learner performance while constraint-based modeling diagnosed errors through violations of domain constraints \citep{AndersonCognitiveTutors,corbett1995KnowledgeTracing,ohlsonConstraintBased}.

Programming tutors extended learner modeling with analysis of the learner's program: early systems estimated mastery over individual programming rules and diagnosed errors from incorrect programs, while later ones used partial program states to generate adaptive feedback \citep{corbett1995KnowledgeTracing,johnson1990Proust,rivers2017datadriven}.
However, such systems generally relied on substantial domain-specific knowledge, making it hard to transfer to a new domain or adapt to an unseen programming problem.

\subsection{Socratic Tutoring}
Socratic tutoring treats questioning as a form of guided inquiry: rather than directly providing the answers, the tutor consistently investigates the learner's assumptions and asks them to reconsider their reasoning \citep{overholser1993socratic,overholser2023socraticReview}. Educational studies suggest that structured Socratic questioning can help critical thinking and conceptual understanding of the material, particularly when learners are required to actively respond to and reflect on the questions posed by the tutor \citep{yang2005socratic,knezic2010socraticDialogue,hsu2022socraticPrompts}. 

Socratic tutors have also been used in the domain of programming education where they guide learners through the program code comprehension by questioning their interpretation of the program behaviour \citep{alshaikh2020socratic,tamang2021comparative}. More recently, LLM-based approaches have extended this tutoring method to multi-turn debugging, investigating the effectiveness of both prompt-based and fine-tuned models as Socratic tutors \citep{alhossami2024socraticdebugging,instructNotAssist23programming,reasoningtrajectoriessocraticdebugging17}. 

However, they all differ in how the learner progress and tutoring objectives are represented and maintained across the conversation, leaving the integration of Socratic interaction with a learned tutoring model in the domain of bug and misconception repair relatively underexplored.

\subsection{LLMs in Programming Education}

Since the emergence of strong LLMs, they have been actively utilised in programming education. Prior work has explored their usefulness in a variety of scenarios, such as code generation, debugging, code explanation, feedback and hint generation \citep{genProgrEducation1progr,newSarsa2022,newRaihan2025LLMCS}. However, the tendency of general-purpose LLMs to provide direct assistance can conflict with the pedagogical objectives that require learners to reason through the problem themselves, while the provided assistance itself may not be related to what the learner would be expected to be taught to solve the programming exercise \citep{codeAid24Progr,instructNotAssist23programming,answerLeakageRobustness11}.

Several approaches have therefore adapted LLM assistance specifically for educational settings. Assignment-specific grounding has been used to keep responses consistent with the course material \citep{tutorsInProgrCourse6}, while pedagogical prompting and guardrails have been used to limit direct solution leakage \citep{codeAid24Progr}.  

Recent works on programming tutors introduce explicit mechanisms for managing the tutoring process through the conversation. TreeInstruct uses multi-turn planning and estimates of the learner's knowledge to determine the next question \citep{instructNotAssist23programming}; TRAVER combines knowledge tracing with turn-level verification to guide learners toward completion of a coding task \citep{traver9Progr}. Misconception-oriented work structures debugging around reasoning trajectories intended to expose faulty beliefs about the program behaviour
\citep{reasoningtrajectoriessocraticdebugging17}. 

Outside of the domain of programming education, ScaffoldLM demonstrates that pedagogical planning, learner state assessment and progress tracking can be incorporated into a fine-tuned tutoring model for teaching mathematics \citep{scaffoldLM19math}. However, the integration of planning-based Socratic tutor that supports bug and misconception repair into a learned tutoring model remains underexplored.

\section{Methodology}
\subsection{SocraticRepair Framework}
\begin{figure}[H]
    \centering
    \includegraphics[
    width=\linewidth
]{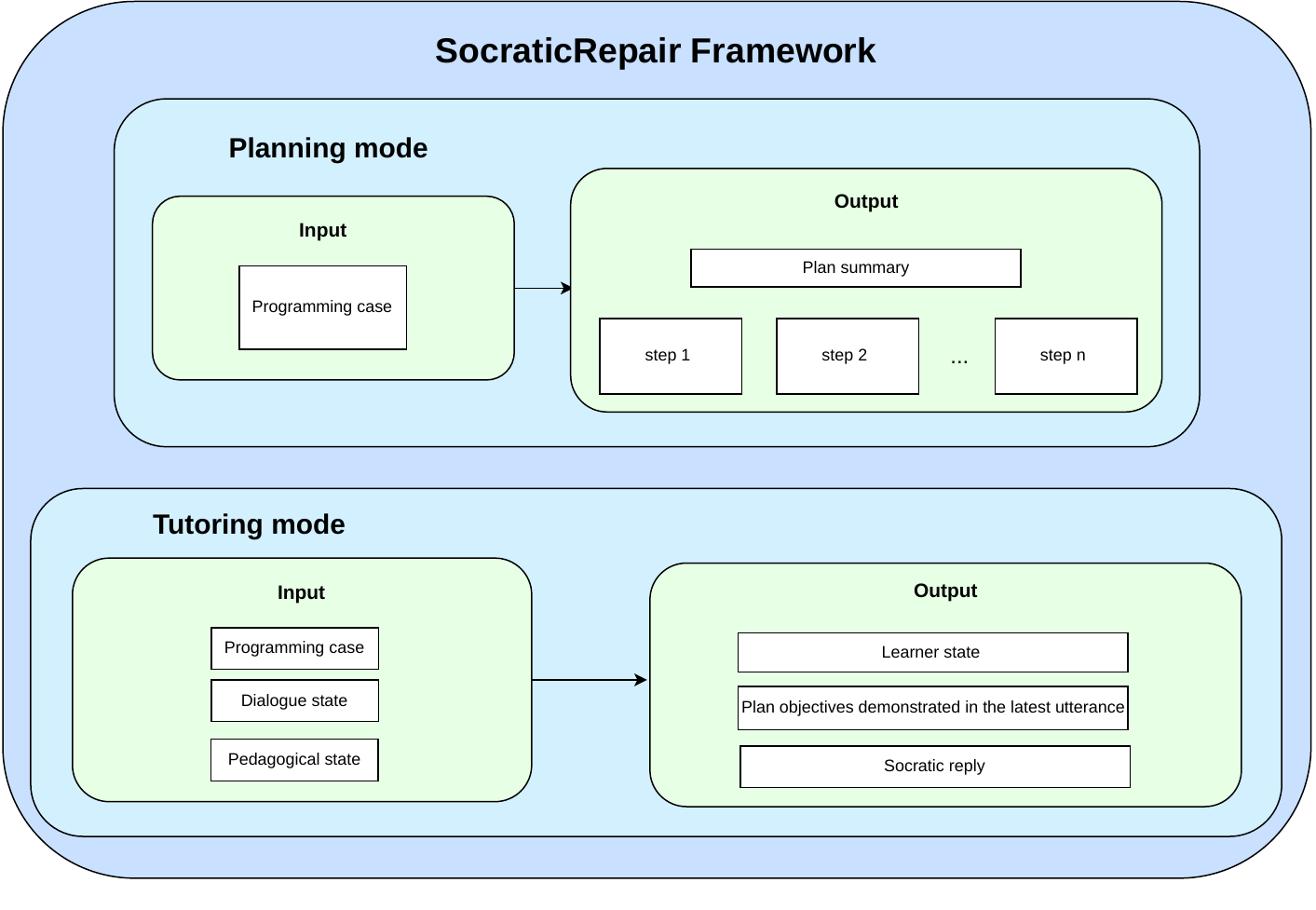}
    \caption{\textbf{SocraticRepair framework.} The same fine-tuned model operates in planning and tutoring modes. In planning mode, the model generates a stepwise pedagogical plan from the programming case. In tutoring mode, the model generates the learner state, step assessments (undemonstrated steps that have been demonstrated in the latest student utterance) and a Socratic reply given a programming case, latest submitted code, execution output, a plan and a progress through the plan. }
    \label{fig:socraticrepair-framework}
\end{figure}

SocraticRepair is a planning-guided multi-turn tutoring framework for bug and misconception repair, instantiated for Python programming language. The framework is shown in Figure~\ref{fig:socraticrepair-framework}. We adopt the tutoring framework introduced by ScaffoldLM \citep{scaffoldLM19math}. Their formulation is adapted to program and misconception repair by defining tutoring progress over both conceptual understanding and code repair by incorporating the state of the learner's program as an additional signal. 
The framework operates in two modes: planning and tutoring. In planning mode, the model receives the programming case together with the correct code solution and generates the pedagogical plan which is then used throughout the subsequent tutoring interaction.  
In tutoring mode, the model is conditioned on the programming case, pedagogical plan, current progress, dialogue history and the learner's latest code and execution result. It jointly predicts the learner state, determines which previously unresolved plan objectives have been demonstrated or not demonstrated in the latest student utterance and generates the next Socratic response. Newly demonstrated plan objectives are then accumulated into the progress state supplied at the next tutor turn.

Progress is represented as a set of demonstrated objectives rather than in a strictly sequential manner, because some cases may involve several intertwined bugs, the repair of which can be performed in a different order by different students. The latest student utterance may therefore demonstrate any currently undemonstrated objective, including objectives scheduled later in the plan, while the first remaining undemonstrated objective is regarded as the current tutoring target.

The learner state characterises the latest student utterance relative to the current plan step and determines the tutor's next pedagogical action. The completion of the plan step is done separately: a correct or comprehension state are only necessary but not sufficient for the demonstration of the current step, since sufficient evidence for the objective must also be present. Table~\ref{tab:learner-state-actions} summarises the corresponding tutoring policy. \\

\begin{table}[H]
\centering

\begin{tabularx}{\textwidth}{@{}lXl@{}}
\toprule
\textbf{Learner state} &
\textbf{Tutoring action} &
\textbf{Active objective} \\
\midrule

START &
Ask an initial guiding question &
Unchanged \\

INCORRECT &
Re-ask or provide a targeted hint &
Not demonstrated \\

CONFUSION &
Simplify the current objective or provide a hint &
Not demonstrated \\

QUESTION &
Answer the clarification and steer the student back to the objective &
Not demonstrated \\

IRRELEVANT &
Refocus the student on the current objective &
Not demonstrated \\

CORRECT / COMPREHENSION &
Advance when sufficient evidence demonstrates the objective, otherwise continue eliciting evidence &
May be demonstrated \\

\bottomrule
\end{tabularx}

\caption{Learner state tutoring policy, adapted from \citep{scaffoldLM19math}. Unlike ScaffoldLM, SocraticRepair requires sufficient evidence that a pedagogical objective has been demonstrated before marking it as completed; the learner state alone is not sufficient.}
\label{tab:learner-state-actions}
\end{table} 
Whenever the learner proposes modified code, it is executed against the available tests. Functional correctness is therefore treated separately from demonstrated conceptual progress: the program code may satisfy the tests while pedagogical plan objectives may still remain unresolved. The interaction therefore terminates only when all the plan steps have been marked as demonstrated and the student's code passes all the tests.

\subsection{Source Datasets and Preprocessing}

The training and evaluation data are constructed from four Python programming datasets: PYMETA \citep{li2026pymeta}, MULTI\_DEBUG \citep{instructNotAssist23programming}, QUIXBUGS \citep{lin2017quixbugs} and INTRO\_PROG \citep{koutcheme2023introprog}. Together, these datasets provide variation in problem complexity and bug structure. 

The datasets differ substantially in their original structure and available information and therefore require dataset-specific preprocessing and conversion to the common benchmark representation. Table~\ref{tab:dataset-preprocessing} records the origin of each of the fields for all the source datasets. Cases for which the required information cannot be established reliably or the case itself is malformed are discarded. \\

\begin{table}[H]
\centering
\setlength{\tabcolsep}{3pt}
\renewcommand{\arraystretch}{1.25}

\begin{tabularx}{\textwidth}{@{}l*{8}{Y}@{}}
\toprule
\textbf{Dataset} &
\makecell{\textbf{Case}\\\textbf{ID}} &
\textbf{Problem} &
\makecell{\textbf{Buggy}\\\textbf{code}} &
\textbf{Tests} &
\textbf{Failure} &
\textbf{Bugs description} &
\makecell{\textbf{Correct}\\\textbf{code}} &
\textbf{Source} \\
\midrule

PYMETA
& D & G & D & G & X & G & G & D \\

MULTI\_DEBUG
& D & D & D & G & X & D & D & D \\

QUIXBUGS
& D & D & D & D & X & G & D & D \\

INTRO\_PROG
& D & C & C & D & X & G & C & D \\

\bottomrule
\end{tabularx}

\caption{Origin of the fields used in the common benchmark representation.
C denotes a value copied directly from the source dataset;
D denotes a value deterministically extracted;
G denotes content generated or reconstructed by an LLM-based preprocessing agent;
and X denotes a value obtained by executing the buggy implementation against the tests.}
\label{tab:dataset-preprocessing}
\end{table}

To reduce leakage between partitions, cases corresponding to the same underlying programming task are grouped before splitting. Candidate groups were constructed using the semantic similarity between programming tasks and the resulting grouping was manually inspected for consistency and potential overlap. Groups were then assigned as indivisible units to the training, validation and test sets.
 Table~\ref{tab:dataset-splits} shows the resulting distribution across the four data sources of 526 programming cases into the final train, validation and test sets.

\begin{table}[h]
\centering

\begin{tabular}{lrrrr}
\hline
\textbf{Dataset} & \textbf{Train} & \textbf{Validation} & \textbf{Test} & \textbf{Total} \\
\hline
PYMETA       & 242 & 29 & 29 & 300 \\
MULTI\_DEBUG & 120 & 15 & 15 & 150 \\
QUIXBUGS     & 32  & 4  & 4  & 40  \\
INTRO\_PROG  & 26  & 5  & 5  & 36  \\
\hline
\textbf{Total} & \textbf{420} & \textbf{53} & \textbf{53} & \textbf{526} \\
\hline
\end{tabular}
\caption{Distribution of programming cases across data sources and train, validation  and test partitions.}
\label{tab:dataset-splits}
\end{table}

\subsection{Synthetic Dialogue Generation}
Before dialogue generation, a pedagogical plan and student profile are generated for each programming case using GPT-5-mini and GPT-5-nano respectively.
The plan is conditioned on the programming case and its bug and repair information (training only) and contains three to eight  pedagogical objectives. A separate verifier agent, GPT-5.6-luna, checks the generated plan is consistent and technically correct, otherwise the plan is regenerated up to 3 times.
The student profile is generated once per case and contains a fixed student background with one to five misconceptions about their program. The student profile is present only during the dialogue generation step as it helps the student LLM keep their erroneous beliefs related to the bugs in the program. Both the plan and the student profile are reused across the three behavioural variants: receptive, uncertain and persistent. 

Dialogue generation proceeds by first generating a student response. The student agent, GPT-5-mini, is provided with all the information about the programming case, including the correct code. For each programming case, 3 dialogues are generated, each with one of the three student variants. The behavioural variant determines a probability distribution over turn-level learner states. The first student turn uses the start state, whereas the subsequent turns are sampled from correct, incorrect, question, comprehension, confusion and irrelevant according to Table~\ref{tab:student-state-probabilities}. This allows to simulate different student behaviours for the same programming problem.
The student verifier agent, GPT-5.6-luna, checks that the student's utterance is consistent with the sampled learner state, otherwise it is regenerated up to 3 times. \\

\begin{table}[H]
\centering
\begin{tabular*}{\textwidth}{@{\extracolsep{\fill}}lccc@{}}
\toprule
\textbf{Learner state} &
\textbf{RECEPTIVE} &
\textbf{UNCERTAIN} &
\textbf{PERSISTENT} \\
\midrule
CORRECT       & 0.40 & 0.30 & 0.25 \\
INCORRECT     & 0.20 & 0.15 & 0.35 \\
QUESTION    & 0.10 & 0.25 & 0.15 \\
COMPREHENSION & 0.10 & 0.10 & 0.05 \\
CONFUSION    & 0.10 & 0.15 & 0.15 \\
IRRELEVANT    & 0.10 & 0.05 & 0.05 \\
\bottomrule
\end{tabular*}
\caption{Learner state sampling probabilities for the three simulated student variants.}
\label{tab:student-state-probabilities}
\end{table}

For each accepted student response, the tutor agent, GPT-5-mini, is provided with all the information about the programming case as well as the plan, current progress, latest submitted code and dialogue history to generate the next Socratic reply. If the student provides the new code, the tutor executes it and the output is included in the prompt. The tutor verifier agent checks that the tutor response is technically correct and aligns with the current active step and the whole dialogue overall, otherwise it is regenerated up to 3 times. 

Dialogue generation continues until all the steps in the progress frontier are marked as completed and the latest student code passes all the tests. If the number of rounds exceeds 14 or the dialogue verifier agent, GPT-5.6-luna, spots an overall inconsistency of the dialogue,  then it is regenerated up to 3 times.

\subsection{Training Data Construction and Fine-tuning}
A total of 1213 accepted dialogues were generated, covering 419 unique programming cases and containing 8493 tutor turns together with 419 pedagogical plans and 419 student profiles. The resulting dialogues and pedagogical plans are converted into two supervised tasks corresponding to two operating modes of SocraticRepair: pedagogical planning and tutoring.
For the planning mode, each example contains the programming case and correct code solution as input, with the pedagogical plan and plan summary as the target.
For the tutor mode, each tutor turn is converted into an individual example conditioned on the programming case, pedagogical plan, current progress, current program state and dialogue history. The target consists of the learner state, a list of demonstrated or not demonstrated decisions for every yet undemonstrated plan step and a Socratic reply.  Inconsistent turn-level samples have been removed, reducing the 8493 generated tutor turns to final 8183 examples. 

The two tasks are combined to fine-tune Qwen2.5-Coder-7B-Instruct using supervised fine-tuning with Low-Rank Adaptation (LoRA). The 419 plans are repeated twice during training and mixed with all 8183 tutor turns before shuffling. Training is performed for two epochs with LoRA rank 32, scaling factor 64 and a dropout 0.05 using a learning rate of \(1 \times 10^{-4}\), an effective batch size of 8 and a maximum sequence length of 12288 tokens. Loss is computed over the assistant completion by masking all prompt tokens.

\section{Evaluation}

\subsection{Evaluation Setup}

We evaluate three tutor variants under the identical controlled student interactions. The first is the complete SocraticRepair pipeline, where the fine-tuned model generates both the pedagogical plan and tutor responses. The second uses the same SocraticRepair tutor with plans generated by the GPT-4o and verified by the GPT-5.6-luna. The third uses GPT-4o which was selected as the general-purpose LLM for Socratic tutoring. The tutors are evaluated through controlled multi-turn dialogues generated on the 53 held-out programming cases. For each programming case, the same student profile is reused across the three behavioural student variants, producing 159 dialogues for each tutor variant. Student responses are generated by GPT-5-mini and verified by the GPT-5.6-luna  agent, with up to three generation attempts. 

The generated dialogues are converted into individual tutor turns where each turn is evaluated in the context of the current programming case, dialogue history and the latest student response. Six turn-level dimensions are adapted from prior tutoring evaluations \citep{scaffoldLM19math,instructNotAssist23programming}. The metrics are summarised in Table~\ref{tab:tutor-quality-metrics}. Premature solution disclosure is recorded separately as a binary turn-level measure, while code repair success and interaction length are reported at the dialogue level.

\begin{table}[H]
\centering

\begin{tabularx}{\textwidth}{@{}lXX@{}}
\toprule
\textbf{Metric} &
\textbf{Evaluated property} &
\textbf{Literature basis} \\
\midrule

Technical correctness &
Technical accuracy and consistency with the program, execution evidence and trusted repair. &
ScaffoldEval: Answer Accuracy \\

Debugging relevance &
Alignment with the student's current objective. &
ScaffoldEval: Topic Adherence; TreeInstruct: Relevance \\

Stepwise scaffolding &
Step by step guidance that leaves meaningful reasoning work to the student. &
ScaffoldEval: Stepwise Scaffolding; TreeInstruct: Indirectness \\

Guidance progression &
Productive movement towards unresolved objectives without unnecessary repetition. &
ScaffoldEval: Guidance Quality; TreeInstruct: Logical Flow \\

Adaptive feedback &
Responsiveness to the student's latest utterance. &
ScaffoldEval: Adaptive Feedback \\

Debugging question quality &
Use of questions that elicit useful explanation &
ScaffoldEval: Question Quality \\

\bottomrule
\end{tabularx}

\caption{Turn level tutor quality dimensions adapted from prior tutoring evaluations \citep{scaffoldLM19math,instructNotAssist23programming}.}
\label{tab:tutor-quality-metrics}
\end{table}

\subsection{Results}

\subsubsection{Dialogue outcomes and premature disclosure}
SocraticRepair successfully completes 148 of the 159 interactions (93.1\%), compared with 152 (95.6\%) for GPT-4o. The 11 incomplete SocraticRepair dialogues consist of eight student generation protocol failures and three tutor output failures, while all seven incomplete GPT-4o dialogues result from student generation failures. \\ SocraticRepair interactions are longer, with a mean of 8.04 rounds compared with 4.99 for the GPT-4o. Premature answer disclosure occurs in 24 of 1,276 SocraticRepair Tutor turns (1.88\%), compared with 59 of 794 GPT-4o turns (7.43\%). The difference is also visible at the dialogue level, where disclosure occurs in 20 SocraticRepair dialogues (12.6\%) and 53 GPT-4o dialogues (33.3\%). This reduction is consistent with SocraticRepair's explicit training for incremental scaffolding and avoidance of premature answer disclosure. Table~\ref{tab:dialogue-results} summarises the main results.

\begin{table}[H]
\centering

\begin{tabular}{lrr}
\toprule
& \textbf{SocraticRepair} & \textbf{GPT-4o} \\
\midrule
Evaluation dialogues & 159 & 159 \\
Successful dialogues & 148 (93.1\%) & 152 (95.6\%) \\
Mean rounds & 8.04 & 4.99 \\
Premature-disclosure turns & 24 / 1,276 (1.88\%) & 59 / 794 (7.43\%) \\
Dialogues with disclosure & 20 / 159 (12.58\%) & 53 / 159 (33.33\%) \\
\bottomrule
\end{tabular}
\caption{Dialogue level outcomes and premature solution disclosure for SocraticRepair and GPT-4o.}
\label{tab:dialogue-results}
\end{table}

\subsubsection{Planner mode quality} 
The learned planner produces a valid pedagogical plan for 52 of the 53 held out cases. Inspection against the trusted programming case information identified a technical error in 41 out of 52 plans, as summarised in Table~\ref{tab:planner-results}. In comparison, all 53 plans generated by the GPT-4o are accepted by the GPT-5.6-luna verifier, with 50 accepted on the first attempt and three after one regeneration. The verified-plan SocraticRepair condition is therefore used to report tutor quality results independently of variation introduced by the plan generation.

\begin{table}[H]
\centering

\begin{tabular}{lrr}
\toprule
\textbf{Plan source} &
\textbf{Attempts} &
\textbf{Technically valid} \\
\midrule
SocraticRepair & 53 & 11 \\
GPT-4o + verifier & 53 & 53 \\
\bottomrule
\end{tabular}
\caption{Technical validity of SocraticRepair-generated pedagogical plans for the held-out evaluation cases.}
\label{tab:planner-results}
\end{table}

\subsubsection{Tutor quality} 
For comparison between tutors, turn-level scores are first averaged within each dialogue. Differences are tested using two-sided Wilcoxon signed-rank tests. Holm correction is applied across the six comparisons to control the family-wise error rate.
Table~\ref{tab:tutor-quality-results} compares the verified-plan SocraticRepair tutor with GPT-4o tutor for the 158 matched dialogues. SocraticRepair achieves higher scores for stepwise scaffolding and debugging question quality, which is consistent with SocraticRepair's fine-tuning objective, whereas GPT-4o obtains higher technical correctness and adaptive feedback scores, which may reflect GPT-4o's stronger general purpose programming capabilities. The difference in scores for the debugging relevance and guidance progression are not statistically significant.

\begin{table}[H]
\centering

\begin{tabular}{lrrrr}
\toprule
\textbf{Metric} &
\textbf{SocraticRepair} &
\textbf{GPT-4o} &
\textbf{$\Delta$} &
\textbf{Holm-adjusted $p$} \\
\midrule
Technical correctness
& 1.789 & 1.860 & -0.071 & .002 \\
Debugging relevance
& 1.929 & 1.949 & -0.021 & .118 \\
Stepwise scaffolding
& 1.876 & 1.789 & +0.087 & .006 \\
Guidance progression
& 1.655 & 1.661 & -0.006 & .385 \\
Adaptive feedback
& 1.792 & 1.843 & -0.051 & .036 \\
Debugging question quality
& 1.626 & 1.446 & +0.180 & $<.001$ \\
\bottomrule
\end{tabular}
\caption{Dialogue averaged tutor quality scores over 158 matched interactions. Scores range from 0 to 2 and $\Delta$ denotes SocraticRepair minus GPT-4o. Statistical significance is assessed using paired two-sided Wilcoxon signed-rank tests with Holm correction across the six metrics.}
\label{tab:tutor-quality-results}
\end{table}

\subsubsection{Adaptation across learner states}
The adaptive feedback metric is at its most important when the student requires a change in the tutoring strategy. Therefore, this metric is examined separately across these learner states: incorrect, question, confusion and irrelevant in Table~\ref{tab:state-conditioned-results}. Across those learner states, neither of the tutors shows a consistent advantage: GPT-4o scores slightly higher on responses following incorrect and question learner states, while SocraticRepair scores higher on responses following confusion and irrelevant. SocraticRepair predicts these four learner states with 91.9--100\% accuracy. Although exact match accuracy for comprehension is only 58.1\%, all 49 misclassifications are predictions of correct, which is still a positive learner state. \\

\begin{table}[H]
\centering

\begin{tabular}{lrrrr}
\toprule
\textbf{Learner state} &
\textbf{Turns (SR/GPT)} &
\textbf{SR Adapt.} &
\textbf{GPT Adapt.} &
\textbf{SR state acc.} \\
\midrule
CORRECT & 430 / 213 & 1.840 & 1.939 & 95.3\% \\
INCORRECT & 236 / 150 & 1.640 & 1.660 & 91.9\% \\
QUESTION & 156 / 103 & 1.833 & 1.864 & 100.0\% \\
COMPREHENSION & 117 / 56 & 1.897 & 1.982 & 58.1\% \\
CONFUSION & 109 / 67 & 1.560 & 1.522 & 98.2\% \\
IRRELEVANT & 69 / 47 & 1.812 & 1.809 & 100.0\% \\
\bottomrule
\end{tabular}
\caption{Adaptive feedback scores by controlled learner state and SocraticRepair learner state prediction accuracy. The fixed \texttt{START} state is omitted.}
\label{tab:state-conditioned-results}
\end{table}

\subsubsection{Performance by bug count} 
Table~\ref{tab:bug-count-results} examines whether tutor behaviour changes with the number of bugs in a programming case. SocraticRepair's advantage in debugging question quality is observed across all of the three groups and is the largest for cases containing three or more bugs. Its technical correctness remains below the GPT-4o one in each group. It can also be observed that even though GPT-4o wins overall on the guidance progression, SocraticRepair scores slightly higher on this metric for cases with three or more bugs, which is consistent with explicit pedagogical planning becoming more useful when several debugging objectives must be tracked. \\

\begin{table}[H]
\centering

\renewcommand{\arraystretch}{1.15}
\begin{tabularx}{\textwidth}{@{}l r *{6}{Y}@{}}
\toprule
\textbf{Bugs} &
\textbf{Matched} &
\textbf{Tech.} &
\textbf{Relevance} &
\textbf{Scaff.} &
\textbf{Progression} &
\textbf{Adapt.} &
\textbf{Q. qual.} \\
\midrule
1  & 74 & -0.094 & -0.004 & +0.137 & -0.015 & -0.037 & +0.157 \\
2  & 42 & -0.057 & -0.045 & +0.014 & -0.017 & -0.098 & +0.113 \\
3+ & 42 & -0.044 & -0.025 & +0.072 & +0.020 & -0.027 & +0.289 \\
\bottomrule
\end{tabularx}
\caption{Difference between SocraticRepair and GPT-4o by annotated bug count ($\Delta=$ SocraticRepair minus GPT-4o).}
\label{tab:bug-count-results}
\end{table}

\subsubsection{Judge agreement}
To assess the reliability of the automatic evaluation, a subset of 40 tutor turns was randomly selected and independently scored by an undergraduate 3rd year Computer Science student using the same metrics. Agreement on the six dimensions is measured using linearly weighted Cohen's $\kappa$, while standard Cohen's $\kappa$ is used for the binary premature disclosure metric. The results in Table~\ref{tab:judge-agreement} demonstrate a substantial agreement across the six tutor quality dimensions. Agreement is highest for the technical correctness, while the more subjective metric of adaptive feedback shows the lowest agreement of 0.66.

\begin{table}[H]
\centering

\begin{tabular}{lr}
\toprule
\textbf{Evaluation dimension} & \textbf{Cohen's $\kappa$} \\
\midrule
Technical correctness        & 0.81 \\
Debugging relevance          & 0.77 \\
Stepwise scaffolding         & 0.70 \\
Guidance progression         & 0.68 \\
Adaptive feedback            & 0.66 \\
Debugging question quality   & 0.72 \\
Premature disclosure         & 0.83 \\
\bottomrule
\end{tabular}
\caption{Agreement between the human annotator and an LLM judge. Linearly weighted Cohen's $\kappa$ is reported for the six tutor quality dimensions, while standard Cohen's $\kappa$ is used for the binary premature disclosure metric.}
\label{tab:judge-agreement}
\end{table}

\section{Conclusion}
This work has introduced SocraticRepair, a planning-guided tutoring framework for multi-turn bug and misconception repair adapted from \citep{scaffoldLM19math}. The framework was instantiated and evaluated in the Python programming domain. The framework combines a pedagogical plan with learner state assessment, progress tracking and executable program feedback. The framework is implemented by fine-tuning Qwen2.5-Coder-7B-Instruct on verified synthetic dialogues constructed from four programming datasets.

The evaluation reveals a clear difference between the reliability of the two learned modes of SocraticRepair. While the tutoring mode demonstrates the intended pedagogical behaviour when conditioned on the verified plans, the learned planning mode frequently produces technically incorrect pedagogical plan steps. Under the verified-plan condition, SocraticRepair tutor mode achieves higher stepwise scaffolding and debugging question quality scores than the GPT-4o and substantially reduces the premature solution disclosure.
GPT-4o remains stronger on technical correctness and adaptive feedback, although learner state conditioned analysis does not show that GPT-4o has a consistent advantage on the learner states where adaptation is most needed.

\section{Limitations and Future Work}

The study has several limitations. The \textbf{first} is that the model is fine-tuned on the synthetic data. While variability in learner behaviour was introduced during the synthetic dialogue generation, these interactions may still not reflect how real learners would interact with the tutor. \textbf{Secondly}, the scale of the synthetic corpus was constrained by the API cost of multi-agent generation, verification and repeated regeneration. Consequently, the model was trained on 1213 accepted dialogues covering 419 programming cases. \textbf{Thirdly}, the learned planning mode remains unreliable, with many generated plans containing technical errors. Future work should therefore investigate the use of a stronger base model with greater programming capability and whether planning should be separated from tutoring into a different agent.
\textbf{Fourthly}, turn-level tutor quality is primarily assessed by an LLM judge. Although human annotation of a subset of turns shows a substantial agreement with the LLM evaluator agent, annotation of all the turn-level turns involving multiple postgraduate human raters would provide more accurate results.

\bibliographystyle{plainnat}
\bibliography{bibliography}

\end{document}